\pdfoutput=1

\documentclass[letterpaper, 10 pt, conference]{ieeeconf}  

\IEEEoverridecommandlockouts                              

\usepackage{graphicx} 

\usepackage[english]{babel}
\usepackage{pgfplots}
\usepackage{pgfplotstable}
\usepackage{booktabs} 
\usepackage{bm} 

\usepackage{tikz}
\usetikzlibrary{shapes,arrows,patterns}
\usepackage{caption}
\tikzstyle{input} = [diamond, draw, 
    text width=4.5em, text badly centered, node distance=3cm, inner sep=0pt]
\tikzstyle{block} = [rectangle, draw, 
    text width=5em, text centered, rounded corners, minimum height=4em]
\tikzstyle{line} = [draw, -latex']

\pgfplotsset{compat=1.11,
    /pgfplots/ybar legend/.style={
    /pgfplots/legend image code/.code={%
       \draw[##1,/tikz/.cd,yshift=-0.25em]
        (0cm,0cm) rectangle (3pt,0.8em);},
   },
   ylabsh/.style={ 
    every axis y label/.style={
      at={(0,0.5)}, xshift=#1, rotate=90
      }
  },
}

\definecolor{colorBluish}{RGB}{101,161,216}
\definecolor{colorBluePastel}{RGB}{139,185,226}
\definecolor{colorGreenish}{RGB}{106,130,94}
\definecolor{colorGreenPastel}{RGB}{150,194,154}
\definecolor{colorRedish}{RGB}{140,21,21}
\definecolor{colorRedPastel}{RGB}{164,117,129}
\definecolor{colorOrangish}{RGB}{237,125,49}
\definecolor{colorOrangePastel}{RGB}{213,169,143}
\definecolor{colorPurplePastel}{RGB}{216,191,216}

\definecolor{colorA}{HTML}{AAAAAA}
\definecolor{colorB}{HTML}{999999}
\definecolor{colorC}{HTML}{777777}
\definecolor{colorD}{HTML}{555555}
\definecolor{colorE}{HTML}{444444}
\definecolor{colorF}{HTML}{222222}
\definecolor{colorG}{HTML}{000000}
\definecolor{colorH}{HTML}{999999}
\definecolor{colorI}{HTML}{777777}

\usepackage{epsfig} 
\usepackage{mathptmx} 
\usepackage{times} 
\usepackage{amsmath} 
\usepackage{amssymb}  

\usepackage{xcolor}
\usepackage{dblfloatfix}    
\usepackage{cleveref} 

\usepackage{float} 

\usepackage{siunitx} 
\usepackage[protrusion=true,expansion=true]{microtype} 

\usepackage[backend=bibtex,style=ieee,maxbibnames=6]{biblatex}

\newcommand{\todo}[1]{\textbf{\textcolor{red}{todo: #1}}}
\renewcommand{\sec}{\second}

\AtEveryBibitem{
\ifentrytype{inproceedings}{
 	\clearlist{address}
 	\clearlist{publisher}
 	\clearname{editor}
 	\clearlist{organization}
 	\clearfield{url}  
 	\clearfield{doi}  
 	\clearfield{pages}  
 	\clearlist{location}
 }{}
 }
 
\AtEveryBibitem{
    \clearfield{url}  
	\clearfield{doi}  
	\clearfield{isbn}
}

\title{\LARGE \bf
Real-time Prediction of Automotive Collision Risk \\ from Monocular Video
}

\author{Derek J. Phillips$^{1\dagger}$ \\ Regina Madigan$^{2}$ \and Juan Carlos Aragon$^{2\dagger}$ \\ Sunil Chintakindi$^{2}$ \and Anjali Roychowdhury$^{3}$ \\ Mykel J. Kochenderfer$^{1}$  
\thanks{*This work is funded by The Allstate Corporation.}
\thanks{$\dagger$Indicates equal contribution.}%
\thanks{$^{1}$Stanford Intelligent Systems Laboratory, Stanford University, Stanford, CA 94305, USA
 		{\tt\small \{djp42, mykel\}@stanford.edu}.}%
\thanks{$^{2}$The Allstate Corporation, Northbrook, IL, USA.}%
\thanks{$^{3}$Department of Mechanical Engineering, Stanford University, Stanford, CA 94305, USA {\tt\small aroyc@stanford.edu}.}%
}

\begin{document}

\maketitle
\thispagestyle{empty}
\pagestyle{empty}

\begin{abstract}
Many automotive applications, such as Advanced Driver Assistance Systems (ADAS) for collision avoidance and warnings, require estimating the future automotive risk of a driving scene.
We present a low-cost system that predicts the collision risk over an intermediate time horizon from a monocular video source, such as a dashboard-mounted camera. 
The modular system includes components for object detection, object tracking, and state estimation. 
We introduce solutions to the object tracking and distance estimation problems.
Advanced approaches to the other tasks are used to produce real-time predictions of the automotive risk for the next \SI{10}{\sec} at over \SI{5}{\hertz}.
The system is designed such that alternative components can be substituted with minimal effort. 
It is demonstrated on common physical hardware, specifically an off-the-shelf gaming laptop and a webcam.
We extend the framework to support absolute speed estimation and more advanced risk estimation techniques.
\end{abstract}

\section{INTRODUCTION}
Predicting automotive risk accurately and efficiently in real time is important for a range of applications.
Potential applications include Advanced Driver Assistance Systems (ADAS), such as collision warning systems, control hand-off guidance for level 3 autonomous vehicles, and planning to avoid high-risk situations for higher-level autonomous vehicles.
Additional applications of real-time automotive risk prediction include variable insurance pricing or driver characterizations for professional drivers.

This work proposes a framework that achieves real-time automotive risk prediction from minimal inputs with reasonable computational power.
The latter two constraints are critical to any system designed for widespread deployment.
Fundamentally, the system is designed to predict the automotive risk that a driver will encounter in the future to a certain time horizon.
This paper uses the term \textit{driver} to refer to the controlling system of the vehicle, which may be either a human or a computer.

Automotive risk can be defined as the ``likelihood and severity of the damage that a vehicle of interest may suffer in the future''~\cite{lefevre:hal}.
In our case, we use time-to-collision (TTC) as a risk surrogate~\cite{blakeAAMAS}, but other risk values can be used instead.
Automotive risk prediction faces many challenges, including partial observability of the scene, a combinatorial explosion of potential vehicle trajectories, and a lack of labeled data with high-risk scenarios for fitting offline models~\cite{blakeAAMAS}.
This paper assumes full observability and develops a system to parse the observations of a single video source to infer a representation of the driving \textit{scene}.
A \textit{scene} is defined to be the state of every vehicle on the road, including position and velocity. 
We use the \textit{scene} to calculate the TTC.

\begin{figure}[t!]
\centering
\scalebox{0.6}{
\begin{tikzpicture}[node distance=2.75cm, auto, thick, every node/.style={scale=0.9, font=\large}]
    \node [input] (image) {Camera};
    \node [input, above of=image, node distance = 2.5cm, dashed] (gps) {GPS};
    \node [block, right of=image, node distance=2.75cm] (obj_det) {Object Detector};
    \node [block, dashed, right of=gps] (speed) {Ego Speed Estimator};
    \node [block, right of=obj_det] (tracker) {Tracker};
    \node [block, node distance=3.7cm, text width=10em, right of=tracker] (traffic) {Traffic Participants' State Estimator};
    \node [block, node distance=3.7cm, right of=traffic] (risk) {Risk Estimator};
    \path [line, dashed] (gps) -- (speed);
    \path [line] (image) -- (obj_det);
    \path [line, dashed] (image) -- (speed);
    \path [line] (obj_det) -- (tracker);
    \path [line] (tracker) -- (traffic);
    \path [line] (traffic) -- (risk);
    \path [line, dashed] (speed) -| (risk);
    \path [dashed] (speed) edge [loop above] node {} (speed);
    \path [] (tracker) edge [loop above] node {} (tracker);
    \path [] (traffic) edge [loop above] node {} (traffic);
\end{tikzpicture}
}
\caption{\small Diagram of the proposed risk prediction system. The dashed components represent extensions of the core system. Self-loop connections represent reliance on stored temporal features.}
\label{fig:high_level}
\end{figure}
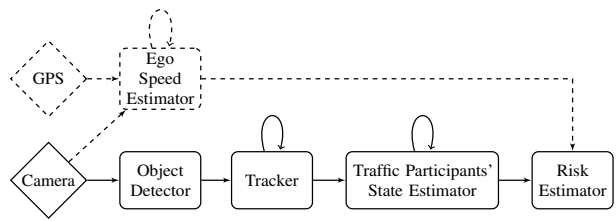

We present a diagram of the proposed risk prediction system in~\cref{fig:high_level}.
The self-loops represent the reliance on stored temporal information.
The dashed components represent optional extensions, which we discuss in~\cref{extensions}.
The system is designed to be modular with interchangeable components for each of the main tasks.
We use deep neural networks for object detection and present new methods for approaching the object tracking and distance estimation.
We propose a particle-based object tracker and a first principles based distance estimation technique, which we show to be efficient and effective. 
These novel approaches complement the deployable and extendable system, which will potentially serve as a  platform for the research community.

We discuss related work in the next section, followed by a problem formulation in~\cref{problem}. 
We then describe our methodology in~\cref{approach} and present relevant results in~\cref{results}.
We do not present an extensive analysis of the risk estimation procedure itself, as that is not the focus of this paper.
In~\cref{extensions}, we present a new solution to the speed estimation problem using detected lane markings as references, which is useful for a variety of applications~\cite{SunilPatent}. 
We conclude by discussing the limitations of our system and future work in~\cref{conclusion}.

\section{RELATED WORK}
\label{rel_work}
To our knowledge, this is the first open-source integrated approach for full-stack automotive risk prediction.
The risk-prediction system described in this paper consists of solving several individual challenges as a system.
Many of these individual challenges have been studied independently.
In this section, we will discuss the relevant work for each of the individual components of the system.

The concept of object tracking has been widely explored and is a critical component in many systems.
Extended object tracking involves tracking multiple objects from frame to frame~\cite{Granstrm2016ExtendedOT}. 
A standard approach is to use information about the location of detected objects to generate probability distributions for their new locations in later frames~\cite{ReidTracking,Granstrom2018LikelihoodBasedDA}. 
Alternative approaches include a two-step solution to first cluster objects and then track them, but we directly use the desired likelihood function by using a sampling approach, which allows for real-time implementation~\cite{Granstrom2018LikelihoodBasedDA,3dtracking}.
A critical component of the tracker is the concept of importance sampling, which is widely used for similar problems~\cite{ImportanceSampling1989,zhaoImportanceSampling}, including particle filters~\cite{kochenderfer2015decision}.

State estimation is a well studied problem with a wide variety of approaches suggested in the literature.
Learning-based approaches~\cite{Saxena_depth_est} and neural networks for depth-estimation~\cite{nips_depth,single_image_depth,Schennings1167554} can encounter issues in highly dynamic environments and in generalizing across cameras~\cite{Schennings1167554}.
Additionally, these approaches can be computationally expensive, and we have limited compute power for our system.
We pursue geometric approaches to estimate distances, and make assumptions about a scene that simplify calculations. 
This approach is has a few similarities to the work of estimating room scenes~\cite{effecient_exact,inside_box}.
Additionally, there are many stereo image based approaches for estimating distance~\cite{CVModern, stereo_depth,stereo_depth2}, which require multiple cameras.

There are many different approaches to predict the risk of a collision, including using simulated collision data based on human driver models to mitigate the lack of data with high risk situations, and learning a prediction model~\cite{blakeAAMAS}. 
Alternatively, one can calculate risk based on an evaluation of potential danger in addition to the probability of a collision~\cite{ammounriskpred}.
Other approaches include the use of a binomial regression model on naturalistic driver data to estimate collision risk~\cite{GUO20133}. 
One of the simplest forms of risk estimation is to use time-to-collision (TTC) of a vehicle.
This technique is widely used, and has been proven to be effective even as a threshold for collision avoidance systems~\cite{netherlandsTTC}.
Due to its simplicity and effectiveness, this paper uses TTC as the risk metric.

\section{PROBLEM}
\label{problem}
The aim of this work is to build an efficient, extendable, and effective system to predict the intermediate-horizon (defined to be \SI{10}{\sec}~\cite{blakeAAMAS}) automotive collision risk for a driver, at low-cost. 
This system will be deployed in a vehicle with a fixed forward-facing camera, such as a dashboard-mounted camera.
For this work, we assume that the parameters of the camera, such as the focal length and mounting position, are known.
We remove expensive sensors such as LIDAR from consideration to minimize the system cost.
Future work includes supporting a full view of the scene through multiple cameras, in order to provide a more complete representation of the scene.

We formulate the automotive risk prediction problem as determining the likelihood of the ego vehicle becoming involved in a collision within \SI{10}{\sec}.
We can define the true parameters of the driving \textit{scene} as $\bm \theta$, which corresponds to all of the information necessary to make an informed prediction of the future.
This information includes the velocities, positions, intentions, and behavioral parameters of all of the traffic participants, including the ego driver.
Our goal in this work is to design a system $g$ that produces an approximation of the true parameters of the scene $\hat{\bm \theta}$. 
A perfect system would produce $\hat{\bm \theta} = \bm{\theta}$. 

We define a risk prediction model $f$ that accepts inputs $\hat{\bm\theta}$ to estimate the probability of a collision occurring over a time horizon $h$ in a scene characterized by $\bm \theta$: 
\begin{align}
  f(\hat{\bm\theta}) \approxeq P(\text{collision}\mid\bm\theta;h)
\end{align}
The system $g$ is designed to output an approximation of the true scene that minimizes $\|\hat{\theta} - \theta\|$.
This approximation must be done in real-time, which we define to be at least at \SI{5}{\hertz}, similar to the human reaction time~\cite{reactionTime}.
This paper focuses on implementing a system $g$ to produce the best approximation of the true parameters of the scene.
We define our time-to-collision based risk model $f$ in~\cref{risk}.

\begin{table}[b]
	\centering
	\caption{Component Interfaces}
	\label{tab:api}
	\begin{center}
		\begin{tabular}{lll} \toprule {\bf Component} & {\bf Input} & {\bf Output} \\
		\midrule
		Camera                  & N/A               & Image (array)\\
		                        &                           & Time in seconds (float) \\
		Object Detector         & Image                     & Object detections \\
		                        &                           & \quad (bounding boxes, labels)  \\
		Tracker                 & Image,                    & Tracked objects \\
		                        & Object detections         & \quad(\{object ID $\to$ \\
		                        &                           & \qquad bounding box\}) \\
		State Estimator         & Tracked objects,          & State info.   \\ 
		                        & Time                      & \quad(\{object ID $\to$ \\
		                        &                           & \qquad rel. distance \& velocity\})\\
		Risk Estimator          & State info.         & Risk score (float)    \\
		\bottomrule
		\end{tabular}
	\end{center}
\end{table}

\section{APPROACH}
\label{approach}
We develop a system that interfaces with different components.
To facilitate system modularity, each component has a defined interface with the other components, and these interfaces are kept to a minimal complexity as shown in~\cref{tab:api}.
Currently, the majority of the system is implemented in Python and uses the Tensorflow~\cite{tensorflow2015-whitepaper} deep learning framework to run the object detection neural network.
Different components perform their tasks on different threads with locked queues serving as interfaces between threads.

\begin{figure}[t!]
    \centering
    \includegraphics[width=\columnwidth]{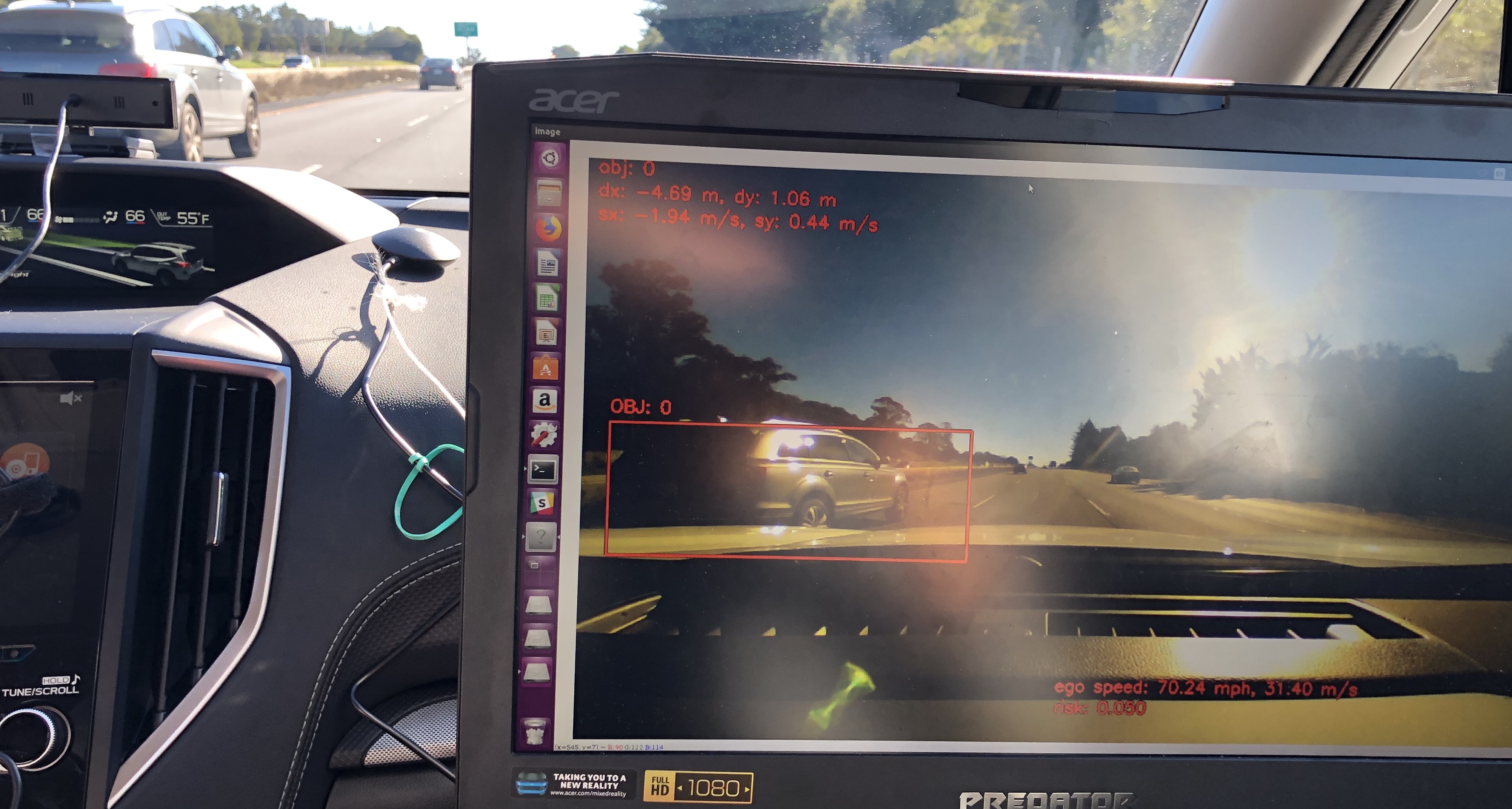}
    \caption{\small The system in use. The webcam is visible in the top left, and the system display on a laptop is shown in the right half of the image. The detected vehicle is labeled object \num{0}.}
    \label{fig:image}
\end{figure}

The system begins by obtaining a raw image from the camera and transmitting it to the object detector.
The object detector outputs bounding boxes directly to the tracker.
These tracked bounding boxes inform the state estimation for the traffic participants.
We define this state as the relative distance and velocity to the ego vehicle. 
The risk estimator uses this state information to calculate the overall automotive risk.
The system is shown in operation in~\cref{fig:image}.

\subsection{Object Detector}
The main functionality of the object detector component of the system is to output bounding boxes and class labels for the objects in a given image. 
A class label is defined to be a key corresponding to the class of an object, such as a ``car'' or a ``person.''
Object detection is a well-known problem, and thus there are many off-the-shelf systems to consider.
Our system uses a FasterRCNN-101 object detection network~\cite{fasterrcnn} that is pre-trained on the KITTI dataset~\cite{kitti}, which focuses on driving data.
The outputs from the network, along with the \textit{frame-time} (time that the image was loaded from the input source) are then added to a queue, which is watched by the subsequent parts of the system. 
To optimize the overall runtime of the system, we dedicate a thread to running the object detector network on the GPU.

\subsection{Object Tracker}
The object tracker incorporates vehicle detections into extended object tracking.
The inputs and outputs of the system are both bounding boxes of vehicles, although the output bounding boxes are also associated with an object ID, which is a key used to identify the object across multiple frames.
Each detected object is assigned a separate object tracker. 
We implement a particle-based tracker with importance sampling~\cite{ImportanceSampling1989}.
We predict the position of a vehicle using particles drawn from a multivariate Gaussian distribution centered on the previous detection of the object.
The object tracker has three main steps: predict, update, and re-sample.

The prediction step samples a new position from the stored distribution.
We add the object's displacement between the last two frames to create the \textit{prior} belief of object position. 
This step is effectively approximating the \textit{motion model} based on the particles from the previous time frame.
This process propagates the $N$ particles forward for an estimate of where the object is in the current frame.
The distribution's variance is set to an appropriate value that accounts for the particular characteristics of vehicle motion.

The update step matches predicted particles with the detections in the frame that are generated by the object detector.
We compute importance weights for all of the particles based on the probability that the given observation would be obtained if the particle's position were the real position.
This update approximates the \textit{perception model}.
The re-sample step generates a new set of particles are drawn from a prescribed distribution based on the weights. 

The resulting distribution of particles after these three steps effectively approximates the \textit{posterior} distribution:
\begin{align}
    P(\mathbf{x}_k\mid \mathbf{z}_{1:k}) =& \notag \\
        \eta P(\mathbf{z}_k\mid \mathbf{x}_k&)
        \int{P(\mathbf{x}_k\mid \mathbf{x}_{k-1})P(\mathbf{x}_{k-1}\mid \mathbf{z}_{1:k-1})
        \mathop{}\!\mathrm{d}\mathbf{x}_{k-1}}
    \label{eq:track1}
\end{align}
Here, $\mathbf{x}_k$ represents the state, which is the object location at time $k$, and $\mathbf{z}_{1:k}$ represents the history of observations. 
The normalization constant $\eta$ accounts for the inverse of $P(\mathbf{z}_k \mid \mathbf{z}_{1:k-1})$.
The \textit{motion model} is $P(\mathbf{x}_k\mid \mathbf{x}_{k-1})$, based on the particles from the previous time frame $k-1$.
We approximate the \textit{perception model} with $P(\mathbf{z}_k\mid \mathbf{x}_k)$, the probability of obtaining observation $\mathbf{z}_k$ if the real object is at location $\mathbf{x}_k$.

The tracker method described above is coupled with a tracker control framework that we design to handle higher level tasks such as tracker initialization and temporary object occlusions while maintaining consistent object-tracker assignments between frames. 
We use heuristics derived from the dynamics of vehicle motion, which makes the tracker particularly effective for vehicle tracking.

\subsection{State Estimator}
The process of estimating the state of other traffic participants is one of the most critical to the successful prediction of automotive risk.
The goal of the state estimator is to determine the relative position and speed of each detected vehicle.
The inputs are the bounding boxes for each detected object, and the outputs are $(d_x, d_y)$ and $(v_x, v_y)$ for each traffic participant.
We define the $x$ axis as lateral motion that is perpendicular to the orientation of the camera, while the $y$ axis is longitudinal motion parallel to the camera.
We make the assumption that the image has been corrected for barrel distortions, which are typical on fish-eye lenses.

\begin{figure*}[t!]
    \centering
    \includegraphics[height=58mm]{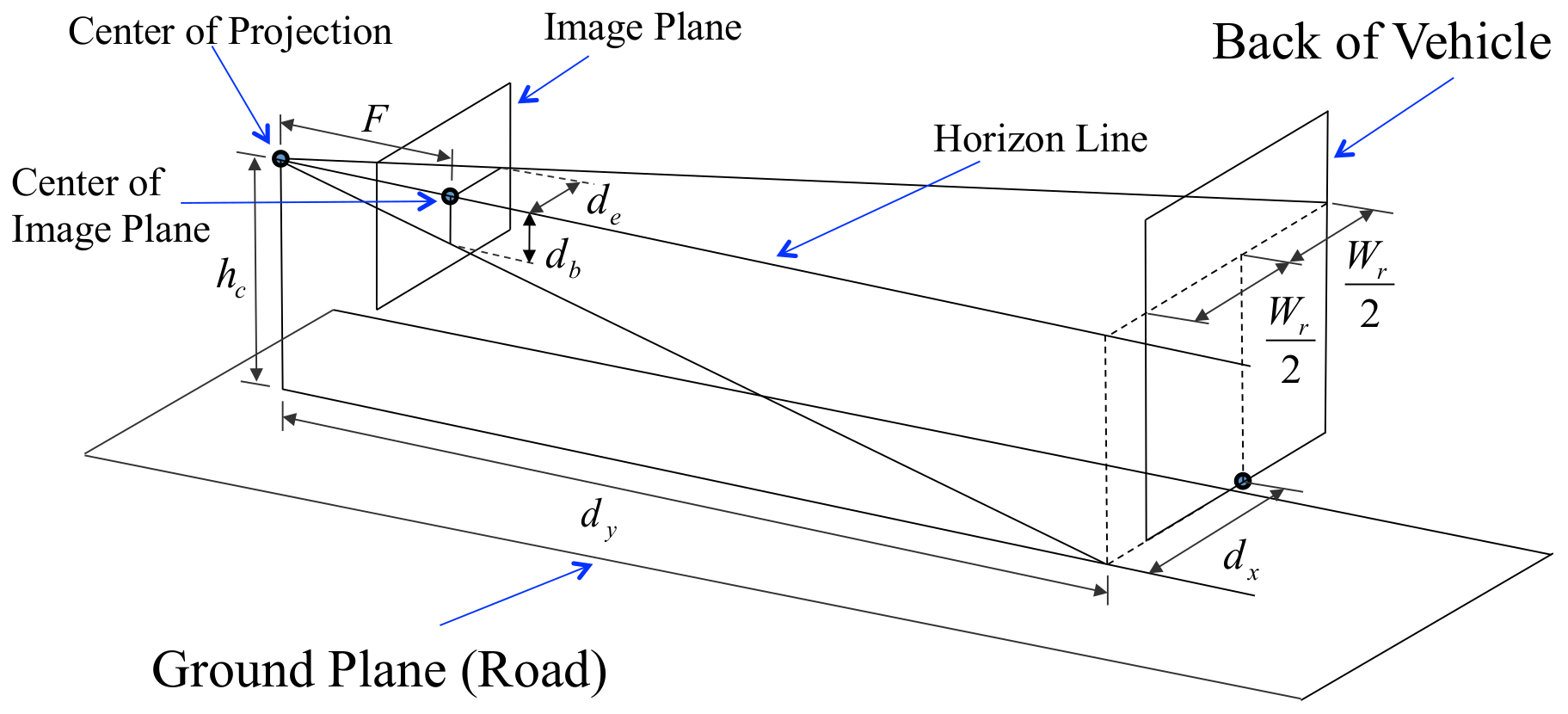}
    \caption{\small The parameters and overall layout of the proposed distance estimation technique.} 
    \label{fig:distance}
\end{figure*}

\subsubsection{Relative Distance}
Our distance calculations are completely geometric in nature, leveraging the following assumptions: we know the camera parameters such as focal length for our input device, and we assume the ground plane is flat and that all other planes we use are orthogonal to the ground plane.
We emphasize that our proposed state estimator is based on a first principles approach, providing both computational advantages and additional robustness over many alternatives.
The only assumption about object dimensions is the width to shift the lateral distance estimate to the vehicle center. 

The main insight is that horizontal lines in the real world are also horizontal in the image frame, in addition to the assumption of a flat ground plane.
We know that the plane of the rear end of the vehicle must intersect with the ground plane (otherwise we would have bigger issues in the scene).
Thus, the distance to the horizontal bottom of the bounding box in the image frame will directly translate to the longitudinal distance to the rear of the vehicle in the real world.

An additional component for the distance estimation process is the image's horizon point.
The horizon point is determined by obtaining the vanishing point of lines that are parallel in the real world, but converge in the image frame due to perspective transformations.
Using the known camera height above the ground $h_c$ and focal length $F$ as above, we can combine this information with distance $d_b$, measured from the bottom of the bounding box in the image to the horizon point:
\begin{align}
  d_y = \frac{h_c F}{d_b}  
\end{align}

We follow a similar process to compute the lateral distance.
We have our estimate of $d_y$, and know the distance from the image center-line to the far edge of the bounding box $d_e$ in pixels.
We use the distance to the far edge of the bounding box because a vehicle may be at a significant angle if it is far off-center, but the far edge will always correspond to the back corner of the vehicle.
As a result, we subtract half the (assumed) car width $W_r$ from the car corner to obtain the lateral distance to the center of the vehicle, $d_x$:
\begin{align}
  d_x = \frac{d_y d_e}{F} - \frac{W_r}{2}  
\end{align}
Due to many constraints, most vehicles have similar exterior width dimensions.
We assume a standard car width of \SI{1.8}{\metre}.

We illustrate the procedure described above in~\cref{fig:distance}.
The camera location is represented by the center of projection and the image plane represents the image frame. 
The image plane is usually referred to as the virtual image plane, since the true image plane is past the center of projection.
If the camera is level (\num{0} tilt angle), the horizon line intercepts the image plane at the center of the image.
Otherwise, the intercept occurs at the vanishing point. 

\subsubsection{Relative Velocity}
The relative velocities are simply the change in relative distance over time. 
Thus, the state estimator component must maintain a history for each of the objects.
At initialization, we assume the relative velocity is \num{0}.
We currently assume low noise in our relative distance estimates, and future work will include advanced filtering techniques to produce smoother estimates.

\subsection{Risk Estimator}
\label{risk}
For the risk model $f$ we use the inverse time-to-collision (TTC) between the controlled (ego) vehicle $e$ and all other detected cars $C$ for which we know $(d_x, d_y, v_x, v_y)$:
\begin{align}
\text{Risk}(\hat{\theta}) &= \max_{c\in C}\frac{1}{\text{TTC}(c,e)}
\end{align}
We predict the TTC by simulating the detected vehicles forward using a constant relative velocity until we reach the maximum time horizon (\SI{10}{\sec}) or encounter a collision.
We define a collision based on assumed vehicle dimensions to determine the minimum acceptable distance between vehicles.
If no detected vehicles are on a collision path, we assign a risk of \num{0}.

\section{RESULTS}
\label{results}
We present the results of experimentation and verification of the components of our system in this section.
All experiments are conducted on commercially available hardware: a gaming laptop with an NVIDIA GTX 1070 GPU, an Intel i7 7700HQ processor, and \num{16}GB of DDR4 memory. 
Input video was provided by a Logitech Genius wide-angle webcam. All tests in this section are conducted on video with a resolution of 1080p, unless otherwise noted, although our system supports any resolution input.
We measure how long the components of the system take to run, as well as the accuracy of specific components.
Runtime experiments involve profiling the system through its execution on a variety of multi-minute collected video segments, totaling approximately \num{1000} frames.

\subsection{Overall Runtime}
\label{sec:overallruntime}
\begin{table}[b]
	\centering
	\caption{Overall Runtime of Risk Prediction System}
	\label{tab:overall_runtime}
	\begin{center}
		\begin{tabular}{lc} \toprule {\bf Component} & {\bf Time per Frame (\SI{}{\milli\sec})} \\
		\midrule
		Object Detector         & 120.7\\
		Tracker                 & 3.2\\
		State Estimator         & 0.3\\
		Risk Estimator          & 0.4\\
		\midrule
		\bf Total (FPS)         & \bf 5.5629 \\
		\bottomrule
		\end{tabular}
	\end{center}
\end{table}
\Cref{tab:overall_runtime} shows the runtime of the system.
The values represent the wall-clock time spent in each component, averaged over the number of frames processed.
We see that the most time intensive component of the system is the object detector, requiring over \SI{120}{\milli\sec} to run on a single image frame. 
The tracker is much faster, at just over \SI{3}{\milli\sec}, and both the state estimator and risk estimator require under \SI{1}{\milli\sec}. 

The discrepancy in~\cref{tab:overall_runtime} between the total FPS of \num{5.56} and the aggregate runtime for each frame is due to time spent in untracked parts of the code, such as thread handling.
When we choose the single-threaded operation of the system, we see the overall FPS increases to \num{6.05} due to this effect.
However, as we see in~\cref{extensions}, a more complex risk estimator requires the threaded version of the system.

\subsection{Tracker Runtime}
The modularity of our system allows us to easily substitute individual components and test the system on the same videos.
We compare our particle tracker with an open source Kernelized Correlation Filter (KCF)  multi-object tracker from OpenCV~\cite{opencv_library}. 
There are many important differences between our particle tracker and the KCF filter that greatly affect performance.
The KCF filter does not require the results of an object detector, except when initializing, which must be done every time we wish to track a new object.

This independence from the object detector creates major limitations for the KCF filter in our application. 
The set of vehicles observed on the road can vary significantly between frames, but the KCF filter can only track vehicles that have already been detected.
\Cref{tab:tracker} compares the runtime of the KCF tracker to the FasterRCNN object detector and our particle-based tracker.
We focus on the runtime and qualitative performance because we do not have ground truth data to compare the tracker performances. 
Evaluating tracker accuracy requires driving video with labeled objects, but we only have access to labeled driving data of images sampled from video to facilitate the training of object detector networks.
Producing such data is left for future work.

\begin{table}[h!]
	\centering
	\caption{Tracker Runtime (\SI{}{\sec})}
	\label{tab:tracker}
	\begin{center}
		\begin{tabular}{cc} \toprule
			{\bf KCF Tracker} & {\bf Object Detector + Particle Tracker} \\
           \midrule 
			0.5010 & 0.1319 + 0.0050 \\
			\bottomrule
		\end{tabular}
	\end{center}
\end{table}

We see that the KCF tracker is significantly slower than our particle-based tracker, even when accounting for the object detector's runtime.
This slowdown is a result of the method used by the KCF tracker for detecting objects based on \textit{regions of interest} and geometric computer vision techniques such as Haar-Like features~\cite{haar}.
It is clear that the particle tracker is superior to the KCF tracker in terms of runtime.
Additionally, due to the fact that the bounding boxes from the particle tracker are produced by the object detection network and not a substitute method as in the KCF tracker, the detections are more accurate and precise.

\subsection{State Estimator Performance}
We conduct a variety of tests to validate our state estimation techniques. 
Our first test consists of a static environment with vehicles at known positions.
For each position, we record the true distance, the relative distance calculated, and the width of the vehicle in the image frame.
The vehicle width in pixels is necessary as a reference for our subsequent test.
We test \num{10} static distances, from \SI{5}{\metre} to \SI{15}{\metre}.
The estimation technique has an average error of \SI[separate-uncertainty = true]{13.32(330)}{\centi\metre}, showing that the distance methods are accurate.

Using the known vehicle widths, the corresponding true distance, and an identical camera setup, we are able to conduct a dynamic test while driving.
In this test, we use linear interpolation between the vehicle widths at known distances to approximate the distance of any detected object. 
We test four different dynamic scenarios, between \SI{10}{\metre} and \SI{15}{\metre}, comparing the estimated distance from our state estimation technique and the approximated true distance based on the vehicle width in pixels. 
We see a wider range of estimation errors in this case, ranging from \SI{2.13}{\centi\metre} to \SI{44.81}{\centi\metre}, with a larger error associated with further distances.
On average, that error amounted to about \num{0.42}\% of the true distance.
This error is influenced by a variety of propagated factors, including the errors at previous steps and bounding box localization error on the vehicle.

\section{EXTENSIONS}
\label{extensions}
\subsection{Ego Speed Estimator}
Many of the potential applications of this system requires knowing the speed of the ego vehicle~\cite{SunilPatent}.
For example, predicting a risk value that includes the energy of a collision is dependent on the absolute speed that the vehicles are traveling. 
To estimate speed, instead of using optical flow techniques~\cite{Indu2011VehicleTA}, we develop a procedure to use lane markings of known dimensions as stationary references. 
Similar to our proposed state estimator, our speed estimator is based on a first principles approach.
We also include support for GPS-based speed estimation.

Our speed estimator uses a few key assumptions: there are visible dashed lane markings, the markings are white and much brighter than their surroundings, and the markings have known dimensions.
These assumptions are not always satisfied, but they hold for highways in the United States. 
The technique involves three steps:
\begin{enumerate}
\item Detect reference lanes and lane markings on the left and the right of the vehicle using the Lane Segment Detection (LSD) algorithm from OpenCV~\cite{opencv_library}.
\item For both the left and the right side, use a vertically fixed point on the reference lane to record the time it takes for a lane marking to appear, disappear, and appear again, $\Delta t$.
\item Use the known distance between lane markings $d$ to calculate the ego vehicle speed, $\frac{d}{\Delta t}$. 
\end{enumerate}

\begin{figure}[t]
    \centering
    \includegraphics[width=.95\columnwidth,trim={0 5cm 0 0},clip]{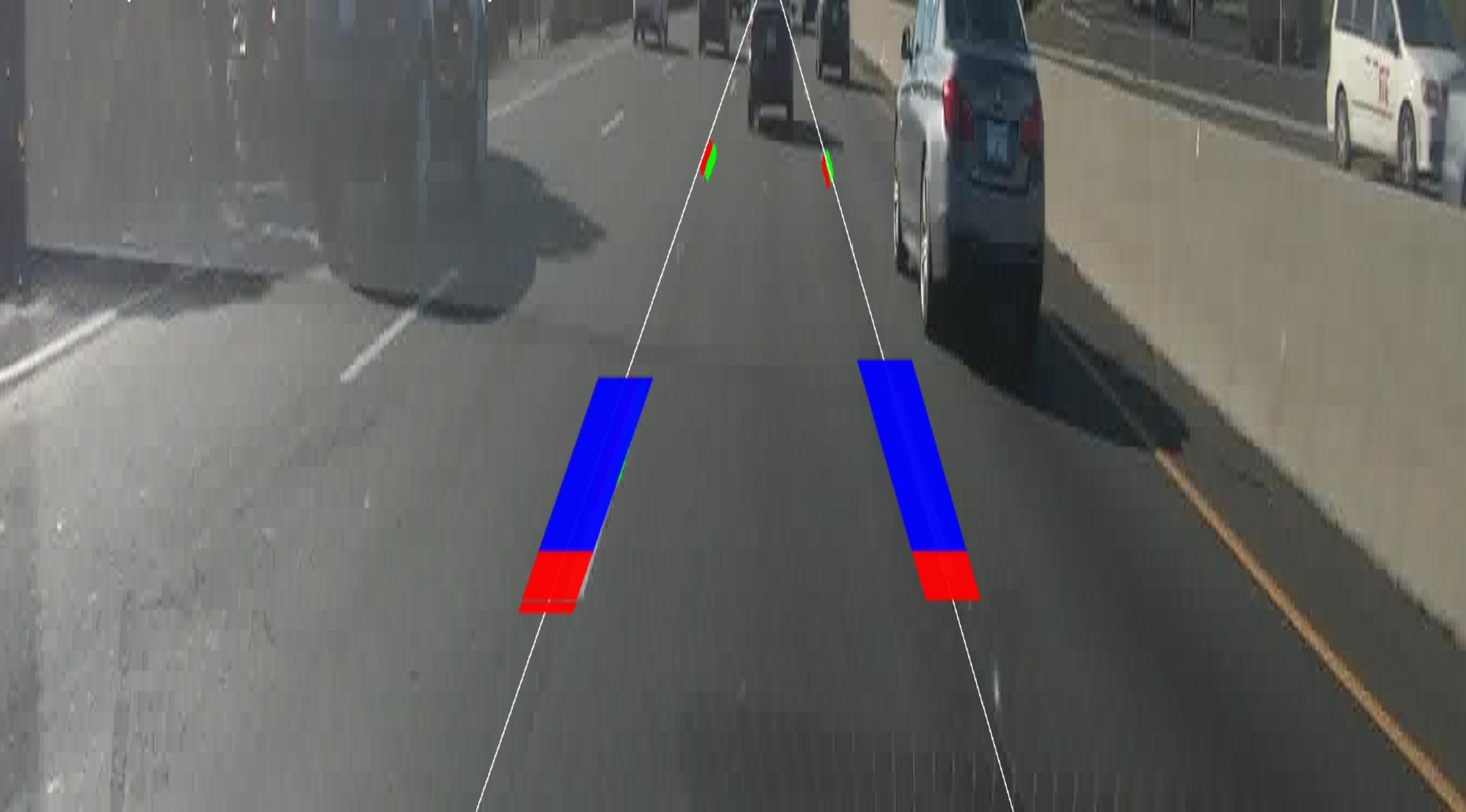}
    \caption{\small The lane based speed estimator uses a fixed vertical position in the image frame to detect and track lane markings over time.}
    \label{fig:speed}
\end{figure}

In~\cref{fig:speed} we show a screen-shot of the lane-based speed estimation technique. 
The road markings that intercept the fixed point of step 2 are highlighted in blue for the part of the marking that is above the fixed point, and in red for the part that is below. 
The left and right referential lines are shown in white superposed over the markings. 
Additional detected white markings have their left borders highlighted in red and their right borders highlighted in green.

We found the lane-based speed estimator predicts a speed approximately \SI[per-mode=symbol]{7}{\kilo\metre\per\hour} below the speed shown on the speedometer of a vehicle when driving at \SI[per-mode=symbol]{65}{\kilo\metre\per\hour}. 
Due to the fact that speedometers in the United States are manufactured to show a speed no more than \SI[per-mode=symbol]{8}{\kilo\metre\per\hour} higher than the actual speed of the vehicle, we consider the speed prediction acceptable.
We saw that the speed estimation is fairly stable over many seconds, and takes \SI{9.31}{\milli\sec} per frame on average.
The technique depends on the dimensions and presence of the lane markings, and requires parameter tuning for different camera resolutions. 

\subsection{Monte Carlo Risk Estimator using Driver Models}
The system presented so far has a relatively simple risk estimation component.
We extend the system by considering an alternative risk estimation technique, inspired by prior work~\cite{blakeAAMAS}.
The idea is to use human driver models to produce rollouts of the traffic participants, resulting in distributed Monte Carlo sampling of future scenes~\cite{TTCP_MC}. 
For each of these scenes, we estimate the risk with the inverse TTC, which is calculated as in~\cref{risk}, and we average the resulting values.
We use the Intelligent Driver Model (IDM)~\cite{idm} for longitudinal acceleration and the MOBIL model~\cite{mobil} for lane change maneuvers, which require the absolute vehicle speed in addition to the relative speed.

We find that this technique results in qualitatively more interesting risk predictions due to the stochasticity of the simulations, whereas the pure TTC risk was much more stable over time.
Overall, the risk estimator requires approximately \SI{106.18}{\milli\sec} per frame, which significantly increases the runtime of the system.
As mentioned in~\cref{sec:overallruntime}, the multi-threaded design of the system is required to make the system run fast enough for our use case.
With the threaded system, we see an overall frequency of about \num{4.9} FPS, while the single-threaded system drops to \num{3.4} FPS.

\section{CONCLUSIONS AND FUTURE WORK}
\label{conclusion}
We have presented a functional real-time system that works with minimal inputs to produce estimates of the future automotive risk of a driving situation.
We propose a particle-filter based object tracking system to track traffic participants, which are identified using a deep neural network object detection system. 
We derive techniques from first principles in order to estimate the relative distance and velocity of the tracked vehicles.
This information is used to predict the expected risk a driver will encounter over the next \SI{10}{\sec} of driving.
The relatively constrained scope and reliance on specific parameters for state and speed estimation are limitations we hope to address in the future.
Our source code is publicly available at~\url{github.com/sisl/MonoRARP}. 

There are many directions that this work can be extended.
One path forward is to incorporate more advanced components into the system, such as Generative Adversarial Imitation Learning~\cite{bhattacharyya2018multi, kuefler2017imitating} to produce more realistic driving scenarios for simulation-based risk prediction.
We will also explore using the system to learn behavioral parameters of our ego driver online and over extended periods of time.
For example, using a given \textit{aggressiveness} parameter, if we continuously predict a risk that is too high, we may adjust the driver's parameter to be more cautious.  
Similarly, we can change parameters that describe driving skill and adjust the risk accordingly to adapt to the specific driver's ability.

We also plan to compare and develop different degrees of end-to-end risk prediction. 
As discussed in previous work~\cite{blakeAAMAS}, risk predictors that are learned offline circumvent the need for simulation and lead to potential performance improvements.
We plan to use an offline version of the system presented in this paper to generate training data for such a system, and compare the resulting performance.

To extend the applicability of the system, we are exploring the use of our state estimation technique on other stationary objects such as lamp posts and street signs in order to produce an estimated ego velocity. 
This will contribute to the approximation of camera parameters from a video, which would hopefully allow us to utilize other sources of data.

\section*{ACKNOWLEDGMENTS}
We thank Anuradha Kodali, Martin Holder, and Blake Wulfe for numerous helpful comments and conversations.


\renewbibmacro{finentry}{%
  \iffieldequalstr{entrykey}{a}
   {\finentry\newpage}
   {\finentry}}
\printbibliography

\end{document}